\documentclass[lineno,twocolumn]{IECEjournal}

\usepackage{graphicx}
\usepackage{amsmath}
\usepackage{amssymb}
\usepackage{enumitem}
\usepackage{tikz}
\usepackage{lipsum}
\usepackage{multicol}
\usepackage{multirow}
\usepackage{booktabs}
\usepackage{tabularx}
\usepackage{lipsum}
\usepackage{algorithm}
\usepackage{algorithmic}
\usepackage{booktabs} 
\usepackage{array}
\usepackage{xcolor} 

\usetikzlibrary{arrows.meta}
\graphicspath{{res/},{media/},{figures/}}

\AtBeginEnvironment{thebibliography}{\small}
\usepackage{newfloat,array}
\newcolumntype{P}[1]{p{#1}<{\raggedright}}
\DeclareFloatingEnvironment[fileext=prc,placement={htb},name=Procedure]{prc}
\newcounter{procedure}
\NewDocumentEnvironment{procedure}{O{ht} m}{%
\begin{prc}[#1]
\caption{#2}
\begin{mdframed}[%
  backgroundcolor=black!10!white,
  font=\ttfamily,
  roundcorner=2pt]
}{\end{mdframed}\end{prc}}

\articletype{Original~Article}
\editor[0000-0001-7897-1673]{Editor A}
\submitdate{submit-date}
\acceptdate{accept-date}
\pubdate{pub-date}  
\jnlsetup{%
    author  = {Xu, D., ... \& Cheng, K. T.},
    journal = {IECE Transactions on Internet of Things},
    title   = {Optimized CNNs for Rapid 3D Point Cloud Object Recognition},
    ctitle  = {and Implementation of 3D Library Application System --- Taking The Library of Henan University of Tech-nology as an Example},
    volume  = {2},
    number  = {1},
    year    = {2022},
    doi     = {10.52810/TPRIS.2021.xxxxxx}
}
\dataavail{%
The data used to support the findings of this study are available from the corresponding author upon request.}

\makeatletter
\def\linenumbershook{}

\makeatother

\begin{document}
\author[1]{Tianyi Lyu}[0000-0000-0000-0000]
\author[2]{Dian Gu}[0000-0000-0000-0000]
\author[3]{Peiyuan Chen}[0000-0000-0000-0000]
\author[4]{Yaoting Jiang}[0000-0000-0000-0000]
\author[5]{Zhenhong Zhang}[0000-0000-0000-0000]
\author[6]{Huadong Pang}[0000-0000-0000-0000]
\author[7]{Li Zhou}[0000-0000-0000-0000]
\author[8][dand97personal@gmail.com]{Yiping Dong}[0000-0000-0000-0000]

\affil[1]{ College of Engineering, Northeastern University, Boston, MA, 02115, United States}
\affil[2]{University of Pennsylvania, Philadelphia, PA, 19104, United States}
\affil[3]{School of Electrical Engineering and Computer Science, Oregon State University, Corvallis, OR, 97333, United States}
\affil[4]{Carnegie Mellon University, College of Engineering, Pittsburgh, PA, 15213, United States}
\affil[5]{George Washington University, Washington, DC, 20052, United States}
\affil[6]{Georgia Institute of Technology, Atlanta, GA, 30332, United States}
\affil[7]{Faculty of Management, McGill University, Montreal, QC, H3B0C7, Canada}
\affil[8]{Department of Mechanical Engineering, Carnegie Mellon University, Pittsburgh, PA, 15213, United States}

\thanks[$\dagger$]{These authors contributed equally to this work}


\maketitle

\abstract{This study introduces a method for efficiently detecting objects within 3D point clouds using convolutional neural networks (CNNs). Our approach adopts a unique feature-centric voting mechanism to construct convolutional layers that capitalize on the typical sparsity observed in input data. We explore the trade-off between accuracy and speed across diverse network architectures and advocate for integrating an $\mathcal{L}_1$ penalty on filter activations to augment sparsity within intermediate layers. This research pioneers the proposal of sparse convolutional layers combined with $\mathcal{L}_1$ regularization to effectively handle large-scale 3D data processing. Our method's efficacy is demonstrated on the MVTec 3D-AD object detection benchmark. The Vote3Deep models, with just three layers, outperform the previous state-of-the-art in both laser-only approaches and combined laser-vision methods. Additionally, they maintain competitive processing speeds. This underscores our approach's capability to substantially enhance detection performance while ensuring computational efficiency suitable for real-time applications.
}

\keywords{Object Detection, $\mathcal{L}_1$ penalty, Point Cloud, MVTec 3D-AD} 


\section{Introduction}

In applications such as autonomous driving and mobile robotics, 3D point cloud data plays a crucial role, and effective object detection is essential for planning and decision-making. While convolutional neural networks (CNNs)~\cite{qiao2024robust,zhu2024complex,liu2024design,wang2023computational,chen2024few,de2022modeling,wang2024deep,peng2024automatic,Shen2024Harnessing,weng2024comprehensive,liu2025eitnet} have recently revolutionized computer vision, especially in 2D tasks (e.g., \cite{krizhevsky2012imagenet}, \cite{simonyan2014very}, \cite{hu2018relation}, \cite{pan20213d}), methods for processing 3D point clouds have yet to experience a similar breakthrough.

The primary computational challenge arises from the third spatial dimension. It is difficult to directly transfer CNNs from 2D visual tasks (e.g., \cite{chauhan2018convolutional}, \cite{fathy1995image}, \cite{liang2017enhancing}) to native 3D perception in point clouds for large-scale applications due to the increased size of input data and intermediate representations. Traditional methods typically involve converting 3D point cloud data into 2D structures, which disrupts the spatial relationships in the original 3D space, requiring the model to reconstruct these geometric correlations.

Moreover, the complexity of 3D data arises not only from larger datasets but also from more intricate spatial dependencies, which standard 2D CNNs are not designed to handle efficiently. This conversion from 3D to 2D may result in a loss of critical spatial information, complicating the learning process as the model must infer 3D structures from 2D projections.

In addition, processing 3D point clouds requires significantly more computational resources, including memory and processing power, which can be a limiting factor for real-time applications like autonomous driving, where fast decision-making is essential. As a result, there is a growing need for specialized architectures and algorithms that can natively handle 3D data, preserving spatial integrity and processing the information efficiently without relying on dimensionality reduction.

In mobile robotics, point cloud data often exhibits spatial sparsity, with many regions remaining unoccupied. This characteristic was effectively exploited in Vote3D, a feature-centric voting algorithm introduced by \cite{wang2015voting}. The algorithm takes advantage of the inherent sparsity in point clouds, scaling its computational cost with the number of occupied cells rather than the total number of cells in the 3D grid.

The research in \cite{wang2015voting} demonstrates that their voting mechanism is equivalent to a dense convolution operation. By discretizing point clouds into 3D grids and performing exhaustive 3D sliding window detection using a linear Support Vector Machine (SVM) \cite{suthaharan2016support}, Vote3D achieved state-of-the-art performance in both accuracy and speed for detecting cars, pedestrians, and cyclists in point clouds.

This effectiveness was validated using the MVTec 3D-AD Vision Benchmark Suite \cite{geiger2012we}. By focusing computational resources only on the occupied cells, Vote3D efficiently handles the sparse nature of point clouds, overcoming one of the key challenges in 3D perception for mobile robotics. This innovation not only improves detection accuracy but also significantly enhances processing speed, making it a pivotal advancement in the field.

Inspired by \cite{wang2015voting}, we propose a novel approach that uses feature-centric voting to directly construct efficient CNNs for object detection in 3D point clouds, without reducing the data to a lower-dimensional space or restricting the search area of the detector. Our method can learn complex, non-linear models and achieve constant evaluation during testing, distinguishing it from non-parametric methods.

To further exploit the computational advantages of sparse inputs throughout the CNN architecture, we introduce an $\mathcal{L}_1$ regularizer during training. This regularizer promotes sparsity not only in the input layer but also in intermediate layers, improving computational efficiency.

Our method fully leverages the sparsity inherent in 3D point clouds, ensuring that computational resources are focused only on occupied regions. This strategy not only improves detection accuracy but also maintains high efficiency, making it ideal for real-time applications in mobile robotics, such as autonomous driving. By processing the 3D data in its native form, our approach preserves the spatial integrity and fine-grained details of point clouds, leading to superior object detection performance. The key contributions of this paper include:

\begin{enumerate}
    \item[1] Development of Efficient Convolutional Layers: We designed convolutional layers optimized for CNN-based point cloud processing, using a voting mechanism to take advantage of the input data's inherent sparsity.
    \item[2] Promoting Sparsity in Intermediate Layers: By incorporating rectified linear units (ReLUs) and applying an $\mathcal{L}_1$ regularization penalty, we ensure sparsity in intermediate representations, which facilitates the use of sparse convolutional layers throughout the entire CNN architecture.
\end{enumerate}

Our experiments demonstrate that our method models perform exceptionally well on the MVTec 3D-AD object detection benchmark within laser-based methodologies. They outperform prior top methods for 3D point cloud-based object detection. This enhancement results in an increase in average precision by up to 40\%. Furthermore, these models maintain competitive detection speeds.

\section{Related Work}
Various studies have investigated the use of convolutional neural networks (CNNs)~\cite{richardson2024reinforcement,huang2024risk,cao2018expected,zhuang2020music,shi2019multi,zhou2024optimization,xu2022dpmpc,peng2024maxk,wang2024intelligent} for processing 3D point cloud data. For example, in \cite{li2016vehicle}, a CNN-based approach is used to achieve comparable results to \cite{wang2015voting} on the MVTec 3D-AD dataset for object detection. This method involves converting the point cloud into a 2D depth map and adding an additional channel to represent the point height above the ground. While this approach allows for the prediction of detection scores and bounding boxes, the conversion of 3D data into a 2D plane leads to a loss of critical information, especially in densely populated scenes. Furthermore, the network is required to learn depth relationships that are naturally embedded in the original 3D data, which could be more effectively captured using sparse convolutions.

Other research, such as \cite{maturana2015voxnet} and \cite{maturana20153d}, has focused on processing dense 3D occupancy grids. For instance, \cite{maturana2015voxnet} reports a GPU processing time of 6ms for classifying a single crop with a grid size of 32×32×32 cells, using a minimum cell size of 0.1m. Similarly, \cite{maturana20153d} reports a processing time of 5ms per cubic meter for landing zone detection. Given that 3D point clouds can cover large areas, such as 60m × 60m × 5m, the total processing time would be approximately 90 seconds per frame (60×60×5×5×${10}^{-3}$), which is impractical for real-time robotics applications.

Additionally, dense grid approaches are computationally expensive and do not scale well with the size of the input data, making them unsuitable for real-time processing in applications like autonomous driving or drone navigation. The need for efficient methods that can handle the high sparsity of 3D point clouds while preserving key spatial information is clear. Our proposed method addresses these challenges by maintaining the full 3D context and leveraging sparsity to reduce computational costs, making it particularly suitable for real-time robotics applications.

Methods utilizing sparse representations were also proposed in \cite{graham2014spatially} and \cite{graham2015sparse}, where sparse convolutions are applied to smaller 2D and 3D crops. However, despite focusing on sparse feature locations, these methods still process neighboring values, which are often zero or constant biases, leading to unnecessary computations and increased memory usage.

Another sparse convolution technique, introduced in \cite{jampani2016learning}, employs "permutohedral lattices." However, this approach is limited to relatively small inputs, unlike our method, which is designed to efficiently handle larger datasets.

CNNs~\cite{wang2024recording,wan2024image,zhang2024deep,wang2021machine,liu2024dsem,zhou2024adapi,jiang2024trajectorytrackingusingfrenet,li2024deep,jiang2020dualvd,weng2024fortifying} have also been applied to process dense 3D data in biomedical imaging, as demonstrated in \cite{chen2016voxresnet}, \cite{dou2016automatic}, and \cite{prasoon2013deep}. For example, \cite{chen2016voxresnet} uses a 3D residual network for brain image segmentation, \cite{dou2016automatic} proposes a two-stage cascaded model for detecting cerebral microbleeds, and \cite{prasoon2013deep} combines three CNNs, each processing a different 2D plane, with the streams merged in the final layer. These systems are primarily designed for smaller inputs and can take over a minute to process a single frame, even with GPU acceleration~\cite{gong2024graphicalstructurallearningrsfmri,liu2025real,de2023performance,ren2024iot,gong2020research,an2023runtime,yan2024application,li2024optimizing,zheng2024identification}.

The need for efficient and scalable methods to process large 3D point clouds remains crucial, particularly for real-time applications in fields like autonomous driving and robotics. Our proposed method addresses these challenges by utilizing sparse convolutions specifically tailored to handle the inherent sparsity of 3D point clouds. This approach reduces computational overhead while preserving the rich spatial information critical for accurate object detection, making it a more practical solution for real-time scenarios.

\begin{figure*}
    \includegraphics[width=1\textwidth]{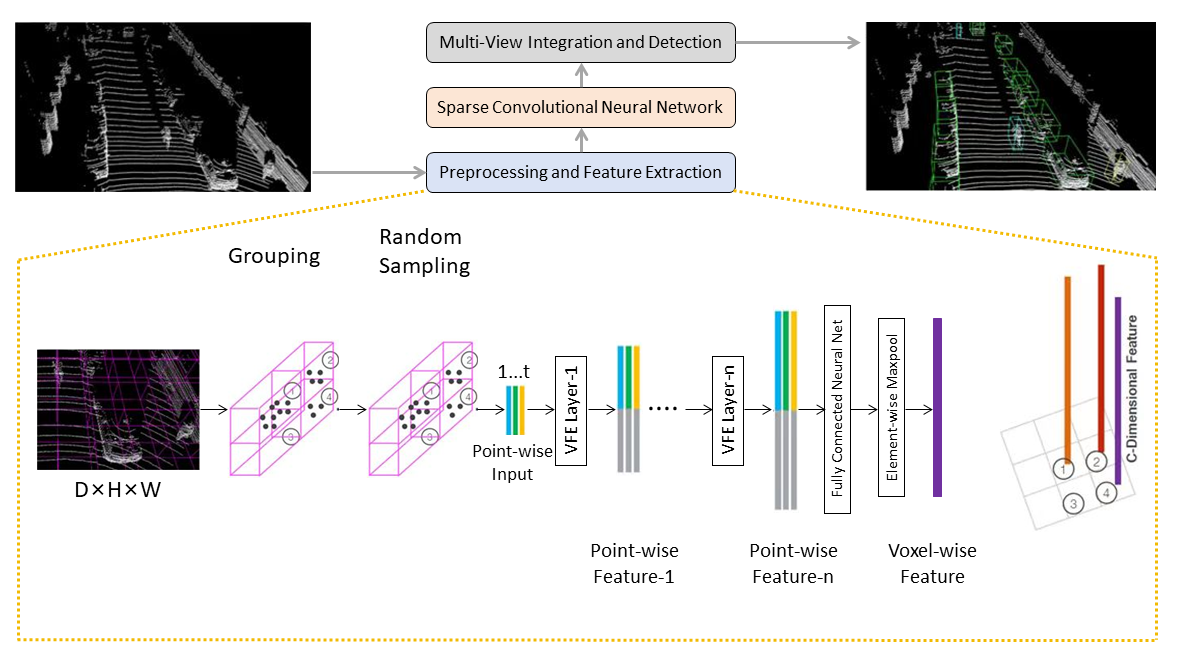}
    \caption{Our architecture.}
    \label{fig2}
\end{figure*}

\section{Our Method}
We propose a method integrating preprocessing, sparse convolutional neural networks~\cite{chen2024enhancing,wang2022classification,lee2024traffic,shen2024deep,wang2018performance,Wang2024Theoretical,chen2024mix,dong2024design,xi2024enhancing}, and multi-view feature integration for efficient 3D point cloud object detection. Preprocessing involves noise reduction and background removal using techniques like RANSAC \cite{derpanis2010overview} and DB-Scan \cite{khan2014dbscan}. Feature extraction combines Fast Point Feature Histograms (FPFH) and multi-view image rendering with ResNet18 \cite{zhou2019facial}. Sparse CNNs~\cite{li2024ltpnet,zheng2024triz,yang2019regional,zhang2024cu,tang2024real,luo2023fleet,sui2024application,xie2023accel,yuan2025gta,wang2024using,wang2024cross}, optimized with importance sampling and hierarchical clustering, enhance computational efficiency. Multi-view integration uses attention mechanisms for robust anomaly detection, ensuring accuracy and efficiency suitable for real-time applications. Our network architecture is shown in Figure \ref{fig2}.

\subsection{Preprocessing and Feature Extraction}
This module prepares the raw 3D point cloud data by removing noise and irrelevant background elements. It also involves extracting meaningful features that capture both geometric and semantic properties of the data. This preprocessing step ensures the efficient and accurate performance of the subsequent modules.

To enhance the quality of the point cloud data, we remove irrelevant background elements. Background elements, such as the ground plane or irrelevant objects, can introduce noise and reduce the efficiency of feature extraction. We use a plane approximation technique that involves selecting a ten-pixel wide strip around the image boundary. By applying RANSAC (Random Sample Consensus) and DB-Scan (Density-Based Spatial Clustering of Applications with Noise) from the Open3D library, we can identify and filter out the background plane.
\begin{equation}
    A_{bg}=\{p \in P \mid distance(p, plane) < \epsilon \}
\end{equation}
This equation defines the background points $A_{bg}$ as those within a certain distance $\epsilon$ from the approximated plane. By removing these points, we obtain a cleaner point cloud 
$P_{clean}$ .

After removing the background, we need to eliminate noise from the point cloud. Noise points can result from various factors such as sensor inaccuracies or environmental interference. We filter out points with NaN (Not a Number) values or those that do not belong to any significant structure. This step ensures that only meaningful data points are retained for further processing:
\begin{equation}
    P_{clean}=P /  A{bg}
\end{equation}
here, $P_{clean}$ represents the cleaned point cloud after removing background points $A_{bg}$.

Fast Point Feature Histograms (FPFH) are used to extract local geometric features from the point cloud. FPFH captures the spatial distribution of points around a given point, providing a detailed representation of the local structure. This is essential for recognizing and distinguishing different objects based on their geometric properties:
\begin{equation}
    F_{FPFH} = {h_1, h_2, ... , h_{33}}
\end{equation}
This equation represents the FPFH feature vector for a point $p_i$, consisting of 33 histogram bins that describe the local geometric properties.

To capture comprehensive information about the object, we generate multi-view images from different angles. This involves rendering the point cloud into 2D images from multiple perspectives. Each view provides a different aspect of the object, enabling the model to learn a more robust representation:
\begin{equation}
    I_v=ender(P, v), v \in V
\end{equation}
here, $I_v$ denotes the image rendered from viewpoint v. The set of viewpoints V is chosen to cover a wide range of angles, ensuring that all significant features of the object are captured.

For the rendered images, we use a pre-trained ResNet18 model~\cite{weng2024big,weng2024leveraging,liu2024td3basedcollisionfree} to extract 2D features. ResNet18 is a deep convolutional neural network that has been trained on the ImageNet dataset, making it capable of extracting high-level semantic features from images:
\begin{equation}
    F_{2D}(I_v)=ResNet18(I_v)
\end{equation}

This equation indicates that the 2D features $F_{2D}$ are obtained by passing the rendered image 
$I_v$ through the ResNet18 model.

Finally, we concatenate the 3D and 2D features to form a comprehensive feature descriptor for each point. This combined feature vector captures both the local geometric information from the point cloud and the high-level semantic information from the multi-view images:
\begin{equation}
    F_{concat}(p_i)={F_{FPFH}(p_i), F_{2D}(I_v)}
\end{equation}

The concatenated feature vector $F_{concat}$  includes both FPFH features and 2D features, providing a rich representation of each point in the point cloud.

\subsection{Sparse Convolutional Neural Networks}
This module introduces the use of sparse convolutional layers to efficiently process the 3D point cloud data. By focusing computations on non-zero elements, we significantly reduce the computational burden while preserving the essential information needed for object detection.

To leverage the spatial relationships between points in the point cloud, we represent the data as a graph G. Each point in the point cloud is treated as a node, and edges are created based on the k-nearest neighbors (k-NN) approach. This ensures that each node is connected to its nearest neighbors, capturing the local structure of the point cloud:
\begin{equation*}
    A_{ij}=\begin{cases}1&\text{if} \space p_j\in\text{k-NN}(p_i)\\0&\text{otherwise}\end{cases}
\end{equation*}

The adjacency matrix A defines the connections between nodes, where $A_{ij}$ is 1 if point 
$p_j$ is among the k-nearest neighbors of point 
$p_i$, and 0 otherwise.

The degree matrix D is calculated as the sum of connections for each node. It represents the number of neighbors each node has and is used to normalize the graph's convolutional operations:
\begin{equation}
    D_{ii}=\sum_j A_{ij}
\end{equation}
here, $D_{ii}$ denotes the degree of node i, which is the sum of the corresponding row in the adjacency matrix A.

Node features are initialized using the concatenated features from Module 1. This provides each node with a rich representation that includes both geometric and semantic information:
\begin{equation}
    H^{(0)}=F_{concat}
\end{equation}
the initial node features $H^{(0)}$  are set to the concatenated feature vectors.

Sparse convolutional layers are applied to propagate information across the graph. These layers focus on non-zero elements, making the computation more efficient. The graph convolution operation is defined as follows:
\begin{equation}
    H^{(l)}=\sigma (D^{-1/2}AD^{-1/2}H^{(l-1)}W^{(l)})
\end{equation}
in this equation, $H^{(l)}$ represents the node features at layer l, A is the adjacency matrix, 
D is the degree matrix, $W^{(l)}$ is the weight matrix for layer l, and $\sigma$ is the activation function (ReLU).

ReLU (Rectified Linear Unit) is used as the activation function to introduce non-linearity into the network. ReLU helps in learning complex patterns by allowing the network to model non-linear relationships:
\begin{equation}
    \sigma(x)=max(0,x)
\end{equation}

This function outputs the input directly if it is positive, otherwise, it outputs zero.

Node embeddings are aggregated to obtain a graph-level representation. This involves pooling the features from all nodes to create a single vector that represents the entire point cloud:
\begin{equation}
    Z=Readout(H^{(L)})
\end{equation}

The Readout operation Z combines the node features $H^{(L)}$ from the final layer to produce a global representation.

Anomaly scores are computed using a multi-layer perceptron (MLP) applied to the graph-level representation. The MLP maps the aggregated features to a score that indicates the likelihood of a point being anomalous:
\begin{equation}
    S=MLP(Z)
\end{equation}

This equation represents the anomaly score S obtained by passing the graph-level representation Z through the MLP.

\subsection{Multi-View Integration and Detection}
This module integrates the features obtained from multiple views and performs anomaly detection by combining the strengths of both 2D and 3D features. This comprehensive approach ensures that the model leverages all available information to detect anomalies accurately.

Features from multiple views are fused to create a robust representation. By averaging the features from different views, we obtain a feature vector that captures information from all perspectives:
\begin{equation}
    F_{fused}(p_i)=\frac{1}{|V|}\sum_{v\in V}F_{2D}(I_v)
\end{equation}
here, $F_{fused}(p_i)$ represents the fused feature vector for point $p_i$, and V is the set of viewpoints.

The combined feature vector includes both 3D geometric information and 2D semantic information. This comprehensive representation is crucial for accurate anomaly detection:
\begin{equation}
    F_{final}(p_i)=Concat(F_{3D}(p_i), F_{fused}(p_i))
\end{equation}

The final feature vector $F_{final}(p_i)$ concatenates the 3D feature $F_{3D}(p_i)$ and the fused 2D features $F_{fuesd}(p_i)$.

The computed anomaly scores are normalized to ensure they are within a comparable range. This step is necessary to standardize the scores across different points:
\begin{equation}
    S_i=\frac{S_i-\min(S)}{\max(S)-\min(S)}
\end{equation}

This normalization formula adjusts the scores $S_i$ to be within the range [0, 1].

A threshold is applied to determine if a point is considered anomalous. Points with scores above the threshold are marked as anomalies:
\begin{equation*}
    \text{Anomaly}(p_i)=\begin{cases}1&\text{if} S_i>\tau\\0&\text{otherwise}\end{cases}
\end{equation*}

This decision rule classifies points as anomalous (1) or normal (0) based on the threshold $\tau$.

Anomalies are localized within the point cloud based on the detection results. The set of anomalous points A is identified by selecting points classified as anomalies:
\begin{equation}
    A={p_i \mid Anomaly(p_i)=1}
\end{equation}

This equation defines the set of anomalous points A as those that meet the anomaly criterion.

\section{Experiments}

This section presents various experiments to assess the performance of ours and highlight the impact of its components on anomaly detection.

\begin{figure*}
    \includegraphics[width=1\textwidth]{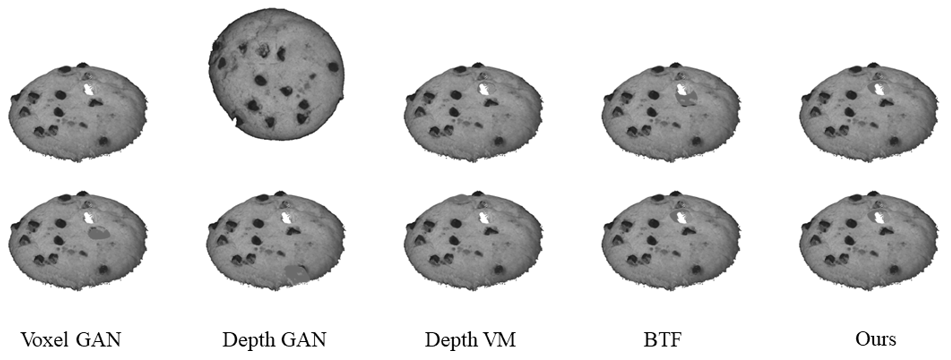}
    \caption{Visualization of prediction results.}
    \label{fig6}
\end{figure*}

\subsection{Experiment Settings} 
\subsubsection{Dataset}
This research examines the MVTec 3D dataset \cite{bergmann2021mvtec}, a newly released real-world multimodal anomaly detection dataset featuring 2D RGB images and 3D PCD scans across ten categories. The dataset encompasses both deformable and rigid objects, some with natural variations (e.g., peach and carrot). Although certain defects are only detectable using RGB data, most anomalies in the MVTec 3D dataset are geometric irregularities. This study primarily investigates PCD anomaly detection, utilizing only the 3D PCD scans in subsequent experiments.

\subsubsection{Implementation Details}
Data Preprocessing: In preparing the point clouds from the MVTec3D dataset, the study first removes irrelevant background elements as outlined in BTF \cite{wei2020view}. This involves using a ten-pixel wide strip around the image boundary to approximate the plane. After eliminating all NaNs (noise) from the PCD, the RANSAC \cite{fischler1981random} and DB-Scan \cite{ester1996density} algorithms from the Open3D library \cite{zhou2018open3d} are employed on this strip to approximate the plane and filter out the background.

3D Modality Feature Extraction: By default, this study adopts the approach used in BTF, utilizing FPFH \cite{rusu2009fast} for extracting 3D modality features. To expedite the computation of FPFH, the point cloud data (PCD) is downsampled prior to feature extraction. The resulting feature dimension for the 3D modality is then calculated accordingly.

2D Modality Feature Extraction: This study generates multi-view images for a given PCD using the Open3D library. The images are rendered at a fixed spatial resolution of 224 × 224. For 2D feature extraction, the first three blocks of ResNet18 \cite{he2016deep}, pre-trained on ImageNet \cite{russakovsky2015imagenet}, are used by default. This process results in a specific feature dimension for the 2D modality.

\subsubsection{Evaluation Metrics}
To evaluate image-level anomaly detection, the area under the receiver operating characteristic curve (AU-ROC) is used, based on the generated anomaly scores. In this study, we refer to this metric as I-ROC for simplicity. For measuring anomaly segmentation performance, the Pixel-level PRO metric (P-PRO) [36] is employed, which accounts for the overlap of connected anomaly components. Following the methodology of previous works \cite{horwitz2023back}, we compute the I-ROC and P-PRO values for each class to facilitate comparison.

\begin{figure*}
    \includegraphics[width=1\textwidth]{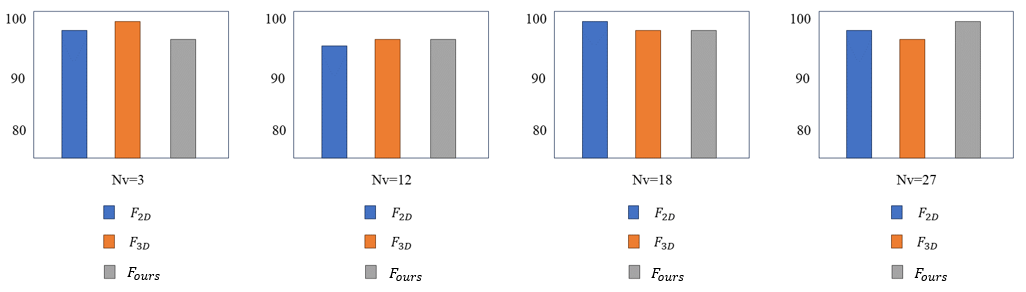}
    \caption{Comparisons of the anomaly detection performances under different types of features, views, and backbones.}
    \label{fig7}
\end{figure*}

\subsection{Comparisons with State-of-the-art Methods}
Table \ref{tab2} and Table \ref{tab3} provide per-class comparisons between Ours and other state-of-the-art methods. These include baselines \cite{bergmann2021mvtec}, AST \cite{rudolph2023asymmetric}, 3D-ST \cite{bergmann2023anomaly}, CPMF\cite{cao2024complementary}, and several benchmarking methods reported in BTF.

In terms of I-ROC, AST currently holds the best average performance among existing methods, achieving an I-ROC of 83.18\%. However, this result still falls short of the optimal performance expected in the field. In contrast, our proposed method, Ours, significantly outperforms all existing techniques, achieving an impressive I-ROC of 95.15\%. Specifically, $F_{Ours}$ excels by securing the highest I-ROC in eight out of ten categories, while ranking second in the remaining two categories—dowel and peach—effectively showcasing the comprehensive superiority of our approach across a wide range of scenarios.

Turning to the P-PRO criterion, which evaluates PCD anomaly localization performance, previous work by BTF demonstrated that handcrafted descriptors were remarkably effective, attaining a notable P-PRO of 92.43\%  using FPFH features. However, our method not only matches this level of performance but surpasses it with a P-PRO of 92.93\%. This improvement underscores our method's enhanced capability in accurately localizing anomalies, which is critical for practical applications in various domains.

Figure \ref{fig6} provides a selection of qualitative results from anomaly detection using Ours, clearly illustrating its effectiveness in identifying geometric abnormalities. These results not only highlight the precision of our method but also demonstrate its robustness across different types of anomalies. By achieving such high performance in both I-ROC and P-PRO metrics, our method sets a new benchmark for future research and applications in anomaly detection.

\subsection{Ablation Studies}
This subsection examines the impact of individual components of Ours, including the number of views for multi-view image rendering, the contributions of 2D and 3D modality features, and the influence of different backbones. Fig. \ref{fig7} compares the performance of Ours across various numbers of views, different feature combinations, and different backbones. Notably, the performance of the 3D modality features remains consistent across all scenarios, as it is determined solely by the 3D handcrafted descriptors used, rather than the number of views or the backbones.

\subsubsection{Influence of the number of rendering views}
Generally, a higher $N_v$ signifies a more comprehensive capture of information. To assess the effect of $N_v$, this study performs multiple experiments with different $N_v$ values, where $N_v \in {1,3,6, ...,27}$. As illustrated in Fig. \ref{fig7}, both I-ROC and P-PRO metrics show moderate improvements as the number of views $(N_v)$ increases, with the most significant rise observed when $N_v$ increases from one to three. There is, however, a slight decline in performance when the number of rendering views is between approximately 12 and 18.

The overall improvements can be clearly attributed to the fact that images from a greater number of views provide a more comprehensive description and capture of the information underlying the given PCD, leading to better performance. For instance, using $N_v$=27 compared to $N_v$=1 brings notable improvements. Specifically, when employing ResNet34 as the backbone for the pre-trained 2D neural networks, there is an approximate increase of 10\% in I-ROC and 4\% in P-PRO when using only 2D modality features $F_{2d}$, and an increase of 5\% in I-ROC and 2\% in P-PRO when using $F_{crmf}$.

The slight drop in performance may be attributed to images from certain specific views generating low-quality features, which can impair anomaly detection. Studies \cite{wei2020view} have shown that adaptive views can better capture the structure of PCD, whereas fixed views might result in poorer performance. Therefore, exploring the selection of optimal views for 2D modality feature extraction could further improve anomaly detection performance.

\begin{figure}
    \includegraphics[width=0.5\textwidth]{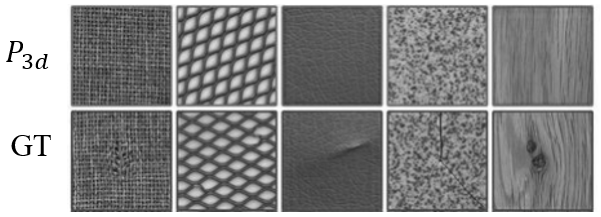}
    \caption{Samples for the influence of different views. }
    \label{fig8}
\end{figure}

Fig. \ref{fig7} displays two examples of images taken from different views. It effectively demonstrates that abnormal regions appear visually distinct depending on the view, and certain views may provide better feature descriptiveness due to clearer imaging of abnormalities. While an increased number of views generally enhances performance, Fig. \ref{fig7} suggests that images from different views contribute differently to anomaly detection. Therefore, selecting views adaptively could significantly improve anomaly detection effectiveness.

\subsubsection{Influence of 2D and 3D modality features}
As mentioned earlier, 3D and 2D feature extraction modules represent PCD data differently, with $F_{3D}$ and $F_{2D}$ containing distinct information. $F_{3D}$ captures extensive geometrical information, whereas $F_{2D}$ focuses on semantics. Fig. \ref{fig7} presents the comparison of anomaly detection performances using various feature combinations. It is evident that using only $F_{3D}$ results in a moderate average I-ROC of 82.04\% and an excellent average P-PRO of 92.30\% across all scenarios. On the other hand, using only $F_{2D}$ shows significant improvement with an increase in the number of views $N_v$.

Specifically, regarding I-ROC, when $N_v$=1, using only $F_{2D}$ does not perform as well as using only $F_{3D}$. However, their combination, $F_{Ours}$, significantly outperforms using either feature type alone. As Nv increases, the performance of using only $F_{2D}$ steadily improves and eventually exceeds that of $F_{3D}$. This suggests that multi-view images can more effectively capture geometrical information in PCD. Moreover, $F_{Ours}$ consistently outperforms single feature types, showing about a 6\% improvement when using ResNet18, effectively highlighting the complementary nature of $F_{2D}$ and $F_{3D}$.

Regarding P-PRO, using only $F_{2D}$ generally results in poorer performance compared to using only $F_{3D}$ across nearly all scenarios, even with the use of multiple views. This may be due to $F_{2D}$ having a larger receptive field than $F_{3D}$, which leads to weaker geometrical point-wise features. The combined feature $F_{Ours}$ slightly outperforms the individual features, showing an improvement of about 0.6\% when ResNet18 is used. In summary, $F_{3D}$ performs better than $F_{2D}$ at the pixel level but worse at the image level. This indicates that the 3D modality features have stronger geometrical information but weaker semantics compared to the 2D modality features. Combining these features provides both local geometrical and global semantic contexts, leading to improved performance at both the image and pixel levels.

\begin{table*}[t]
\centering
\caption{QUANTITATIVE RESULTS (I-AUC). }
\resizebox{\linewidth}{!}{
\begin{tabular}{cccccccccccc}
\hline
Method& Bagel& Cable Gland& Carrot&Cookie&Dowel& Foam&Peach&Potato& Rope&Tire&Mean\\
\hline
 Voxel GAN& 0.3830& 0.6230& 0.4740& 0.6390& 0.5640& 0.4090& 0.6170& 0.4270& 0.6630& 0.5770& 0.5376\\
 
  Voxel AE& 0.6930& 0.4250& 0.5150& 0.7900& 0.4940& 0.5580& 0.5370& 0.4840& 0.6390& 0.5830& 0.5718\\
  Voxel VM& 0.7500& 0.7470& 0.6130& 0.7380& 0.8230& 0.6930& 0.6790& 0.6520& 0.6090& 0.6900& 0.6994\\
 Depth GAN& 0.5300&  0.3760& 0.6070& 0.6030& 0.4970& 0.4840& 0.5950& 0.4890& 0.5360& 0.5210&  0.5238\\
  Depth AE& 0.4680& 0.7310& 0.4970& 0.6730& 0.5340& 0.4170& 0.4850& 0.5490& 0.5640& 0.5460& 0.5464\\
 Depth VM& 0.5100&  0.5420& 0.4690& 0.5760& 0.6090& 0.6990& 0.4500& 0.4190& 0.6680& 0.5200& 0.5462\\
 AST& 0.8810& 0.5760& 0.9560& 0.9570& 0.6790& 0.7970& 0.9800& 0.9150& 0.9560& 0.6110& 0.8318\\
 BTF (Depth iNet)& 0.6860& 0.5320& 0.7690& 0.8530& 0.8570& 0.5110& 0.5730& 0.6200& 0.7580& 0.5900&  0.6749\\
 BTF (Raw) & 0.6270& 0.5060& 0.5990& 0.6540& 0.5730& 0.5310& 0.5310& 0.6110& 0.4120& 0.6780&  0.5722\\
 BTF (HoG)& 0.4870& 0.5880& 0.6900& 0.5460& 0.6430& 0.5930& 0.5160& 0.5840& 0.5060& 0.4290&  0.5582\\
 BTF (SIFT)& 0.7110& 0.6560& 0.8920& 0.7540& 0.8280& 0.6860& 0.6220& 0.7540& 0.7670& 0.5980&  0.7268\\
 CPMF & 0.9812 & 0.8888 & 0.9872 & 0.99892 & 0.9556 & 0.8073 & 0.9856 & 0.9534&0.9781 &0.9678 & 0.9502 \\
 \textbf{Ours}& \textbf{0.9830}& \textbf{0.8894}& \textbf{0.9885}& \textbf{0.9910}& \textbf{0.9578}& \textbf{0.8094}& \textbf{0.9884}& \textbf{0.9590}& \textbf{0.9792}& \textbf{0.9692}&  \textbf{0.9515}\\
\hline
\end{tabular}
}
\label{tab2}
\end{table*}

\begin{table*}[t]
\centering
\caption{QUANTITATIVE RESULTS (P-PRO).}
\resizebox{\linewidth}{!}{
\begin{tabular}{cccccccccccc}
\hline
Method& Bagel& Cable Gland& Carrot&Cookie&Dowel& Foam&Peach&Potato& Rope&Tire&Mean\\
\hline
 Voxel GAN& 0.4400& 0.4530& 0.8250& 0.7550& 0.7820& 0.6970& 0.3780& 0.3920& 0.7750& 0.3890& 0.5828\\
 
  Voxel AE& 0.2600& 0.3410& 0.5810& 0.3510& 0.5020& 0.6580& 0.2340& 0.3510& 0.0150& 0.1850& 0.3478\\
  Voxel VM& 0.4530& 0.3430& 0.5210& 0.6970& 0.6800& 0.6160& 0.2840& 0.3490& 0.6160& 0.3460& 0.4923\\
 Depth GAN& 0.1110&  0.0720& 0.2120& 0.1740& 0.1600& 0.3850& 0.1280& 0.0030& 0.4460& 0.0750&  0.1423\\
  Depth AE& 0.1470& 0.0690& 0.2930& 0.1740& 0.2070& 0.4170& 0.1810& 0.5490& 0.5450& 0.1420& 0.2031\\
 Depth VM& 0.2800&  0.3740& 0.2430& 0.5260& 0.4850& 0.6990& 0.3140& 0.4190& 0.5430& 0.3850& 0.3737\\
 AST& 0.9500& 0.4830& 0.9793& 0.8681& 0.9050& 0.7970& 0.6320& 0.1640& 0.9610& 0.5420& 0.8328\\
 BTF (Depth iNet)& 0.7690& 0.6640& 0.8870& 0.8800& 0.8640& 0.5110& 0.2690& 0.1990& 0.8520& 0.6240&  0.7550\\
 BTF (Raw) & 0.4010& 0.3110& 0.6380& 0.4980& 0.2500& 0.5430& 0.2540& 0.9350& 0.8080& 0.2010&  0.4418\\
 BTF (HoG)& 0.7110& 0.7630& 0.9310& 0.4970& 0.8330& 0.5930& 0.5020& 0.8760& 0.9160& 0.8580&  0.7702\\
 BTF (SIFT)& 0.9420& 0.8420& 0.9740& 0.8960& 0.8974& 0.6860& 0.7230& 0.5270& 0.9530& 0.9290&  0.9094\\
 CPMF &0.9570 &0.9432 &0.9834 &0.9202 &0.9088 &0.9082 &0.7452 &0.9412 &0.9723 &0.9770&0.9282 \\
 \textbf{Ours}& \textbf{0.9730}& \textbf{0.9456}& \textbf{0.9860}& \textbf{0.9210}& \textbf{0.9100}& \textbf{0.9094}& \textbf{0.7460}& \textbf{0.9440}& \textbf{0.9760}& \textbf{0.9773}&  \textbf{0.9293}\\
\hline
\end{tabular}
}
\label{tab3}
\end{table*}

\begin{figure}
    \includegraphics[width=0.5\textwidth]{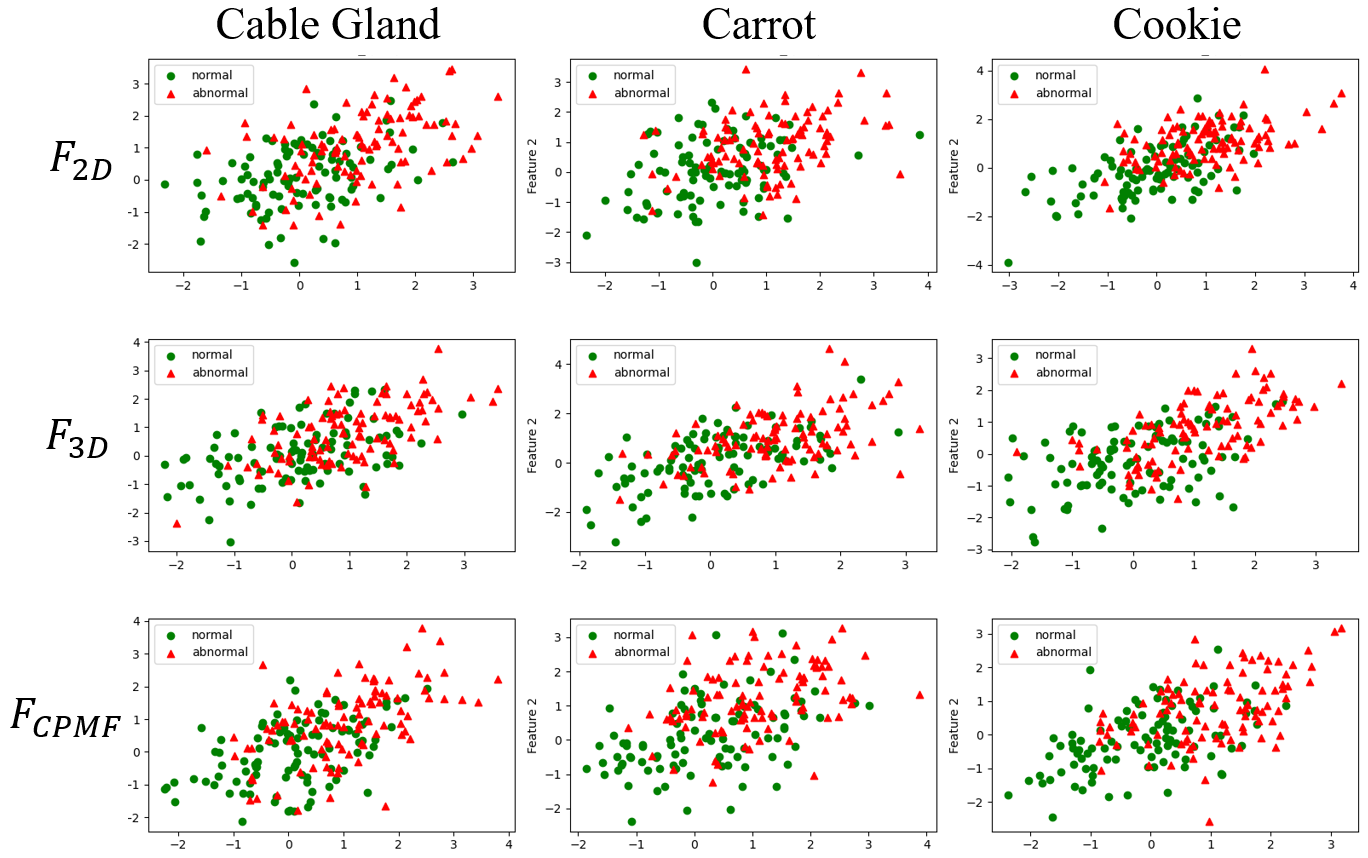}
    \caption{Visualization for feature distributions. }
    \label{fig9}
\end{figure}

Fig. \ref{fig6} demonstrates that $F_{2D}$ and $F_{3D}$ exhibit different strengths in detecting PCD anomalies. For instance, $F_{2D}$ alone effectively localizes anomalies in categories such as cable gland, carrot, and dowel but performs poorly in categories like bagel and potato, where $F_{3D}$ excels. The combination of $F_{2D}$ and $F_{3D}$, forming $F_{Ours}$, markedly enhances anomaly detection performance and achieves impressive localization across all categories, as illustrated in Fig. \ref{fig6}. Fig. \ref{fig9} visualizes the feature distributions of $F_{2D}$, $F_{3D}$, and $F_{Ours}$. It reveals that single-type features may not be well distinguished, whereas the distribution of $F_{Ours}$ is more distinct.

\subsubsection{Influence of backbones}
Fig. \ref{fig7} compares Ours's performance using various backbones, including ResNet18, ResNet34, ResNet50, and Wide\_ResNet\_50\_2 \cite{zagoruyko2016wide}. Table \ref{tab4} summarizes the best performance results for each backbone. Across different backbone types, using only $F_{2D}$ yields poorer pixel-level performance but better image-level performance compared to using only $F_{3D}$. Combining both features enhances performance at both image and pixel levels. Additionally, the backbone type does not significantly impact overall performance, with Ours achieving the best results using ResNet18, boasting a 95.15\% I-ROC and a 92.93\% P-PRO.

\subsubsection{limitation}
\begin{figure}
    \includegraphics[width=0.5\textwidth]{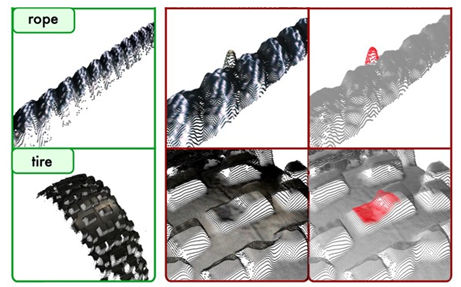}
    \caption{Illustration for the bad quality of rendered images resulting by 
acquisition noise. }
    \label{fig10}
\end{figure}

In this subsection, several limitations are discussed. First, as shown in Fig. \ref{fig10}, the quality of rendered images can be compromised due to noise introduced during PCD acquisition. This degradation in quality can negatively impact feature capability and lead to incorrect judgments. Second, while the current pixel-wise criterion P-PRO effectively reveals the performance of detecting various anomalies, certain anomalies can only be identified with RGB information. This discrepancy results in Ours achieving excellent but not optimal performance and underscores the need for a more equitable metric for point-wise PCD anomaly localization.

\begin{table}[t]
\centering
\caption{QUANTITATIVE RESULTS.}
\resizebox{\linewidth}{!}{
\begin{tabular}{cccc}
\hline
Backbone& Feature&I-ROC& P-PRO\\
\hline
 & P-PRO& 0.8304& 0.9230\\
 \hline
  \multirow{2}{*}{ResNet18}&  873±234& 0.8918& 0.9145\\
  
  & 819±211& 0.9515& 0.9293\\
  \hline
 \multirow{2}{*}{ResNet34}& 814±213& 0.8987& 0.9135\\
  & \textbf{553±134}& 0.9492& 0.9286\\
  \hline
  \multirow{2}{*}{ResNet50}& & 0.8932& 0.8977\\
  & & 0.9479& 0.9233\\
  \hline
  \multirow{2}{*}{Wide\_ResNet\_50\_2}& & 0.8911& 0.9038\\
  & & 0.9464& 0.9256\\
\hline
\end{tabular}
}
\label{tab4}
\end{table}

\section{Conclusion}
This study introduces swift object detection in point clouds by employing CNNs built from sparse convolutional layers, adopting the voting mechanism outlined in \cite{wang2015voting}. Leveraging hierarchical representations and non-linear decision boundaries, our approach attains cutting-edge performance on the MVTec 3D-AD benchmark for point cloud object detection. Moreover, our method surpasses the majority of methods that combine information from both point clouds and images across diverse test scenarios.

Future directions for this research include exploring more granular input representations and developing a GPU implementation of the voting algorithm to further enhance detection speed and efficiency. These improvements could provide even faster and more accurate object detection capabilities in 3D environments, making the approach more viable for real-time applications in autonomous driving and robotics.



\conflictsofinterest
The authors declare that they have no conflicts of interest.

\acknowledgement
This work was supported without any funding.


\begin{fullwidth}
\bibliographystyle{plain}
\bibliography{ref}
\end{fullwidth}

\makeatletter

\makeatother

\end{document}